\documentclass[letterpaper, 10 pt, conference]{ieeeconf}  

\IEEEoverridecommandlockouts                              

\usepackage{graphicx} 
\usepackage{amsmath} 
\usepackage{amssymb}  
\usepackage{multicol}
\usepackage{url}

\usepackage{color}

\title{\LARGE \bf Projection based whole body motion planning for legged robots }

\author{Diego Pardo, Michael Neunert, Alexander W. Winkler and Jonas Buchli%
\thanks{This research has been funded through a Swiss National Science
Foundation Professorship award to Jonas Buchli and by the Swiss National Centre
of Competence in Research Robotics (NCCR Robotics) - Diego Pardo {\tt\small \{depardo@ethz.ch\}},
Michael Neunert {\tt\small \{neunertm@ethz.ch\}}, Alexander Winkler {\tt\small
\{winklera@ethz.ch\}} and Jonas Buchli {\tt\small \{buchlij@ethz.ch\}} are with
the Agile \& Dexterous Robotics Lab at the Institute of Robotics and Intelligent
Systems, ETHZ Z\"urich, Switzerland. }%
}

\begin{document}

\maketitle \thispagestyle{empty} \pagestyle{empty}

\begin{abstract}

In this paper we present a new approach for dynamic motion planning for legged
robots. We formulate a trajectory optimization problem based on a compact
form of the robot dynamics. Such a form is obtained by projecting
the rigid body dynamics onto the null space of the Constraint Jacobian.
As consequence of the projection, contact forces are removed from the model but
their effects are still taken into account. 
This approach allows to solve the optimal control problem of a floating base
constrained multibody system while avoiding the use of an explicit contact model. %
%
%
We use direct transcription to numerically solve the optimization.
As the contact forces are not part of the decision variables 
the size of the resultant discrete mathematical program is reduced 
%
and therefore solutions can be obtained in a tractable time. Using a predefined
sequence of contact configurations (phases), our approach solves
motions where contact switches occur. Transitions between phases are
automatically resolved without using a model for switching
dynamics. We present results on a hydraulic quadruped robot (HyQ), including
single phase (standing, crouching) as well as multiple phase (rearing, diagonal leg balancing and stepping) dynamic motions. \end{abstract}


\section{INTRODUCTION}

Motor skills on state of the art legged robots are still far from mature, as 
they clearly do not move as smoothly and efficiently as animals yet.
Regardless of specific and carefully designed implementations, 
there is no general approach for the online planning and execution of
dynamic whole body motions on legged robots. 
In such systems, using hand designed or motion captured references tends to
produce inefficient and artificial results. %


Other approaches \cite{Winkler2015} use the well known Zero Moment Point (ZMP) \cite{Kajita2003} to
maintain balance during motions. The typical
simplified model used to calculate the ZMP for a legged robot is the cart-table
model, which does not capture the full rigid body dynamics of the physical robot
and can lead to instability. Additionally, the quasi-static stability criterion
requires to keep the ZMP inside the support polygon created by the legs in
contact. While this approach works well in practice, it is based
on several assumptions, 
hindering dynamic and more natural looking motions which would be
feasible from a physics perspective.

On the other hand, whole body motion planning and control 
based on optimization is demonstrating to be a very promising approach
\cite{Dai2014,Posa2014,Koch2014,Mordatch2012}. It is clear that contact forces play a decisive role in the
resultant behavior of the robot, and therefore they need to be taken into
account during the planning process. Trajectory optimization offers a framework
where contact forces can be consistently included as decision variables.
However, a continuous optimal control problem for floating base robots
subject to kinematic constraints is so  often too complex for analytical approaches
and therefore numerical methods must be used.
The choice of numerical technique has a significant
influence on the computational effort and feasibility of the solution.

Numerical methods for trajectory optimization can be roughly classified as
shooting or direct methods and both approaches have been applied to create
dynamic motions for legged robots. In essence, all methods search for a set of
decision variables (e.g., joint torques and external forces) such that a
cost function is minimized while
satisfying a set of constraints (e.g., system dynamics and kinematic
constraints). The minimization is accomplished by iteratively
following a numerical approximation of a gradient.

Broadly speaking, \textit{shooting methods} obtain the data required 
for the gradient approximation by forward simulating the system dynamics and
computing the resulting cost function. Importantly, these methods needs to
include the constraints as penalty terms in the cost function, i.e. adding the
requisite of dynamic feasibility as a soft constraint. As a consequence, these
methods may provide suboptimal solutions or even violate system dynamics. Most
of the results have been verified in animation \cite{Mordatch2012}, i.e., playing back the
state trajectories instead of forward integrating the
equations of motion or applying the planned torques on a robot.
To the best of our knowledge, there are no reports of experiments on torque 
control legged robots using shooting methods.

On the other hand, \textit{direct methods} discretize the state, control and
force trajectories over time and an optimization problem is stated over the
values of the variables at the discretization points or nodes \cite{Posa2014,Dai2014}.
The resulting problem can then be solved by a nonlinear
programming (NLP) solver. One of the main advantages of direct methods is that
constraints are enforced by the 
solver. This feature facilitates applying solutions to real robots, for
instance adding bounds on the state and control variables. 

As a consequence of
discretizing both state and control, the dynamics of the system have to be
verified in between the nodes, which increases the number of constraints.
The large size of the subsequent discrete optimization is an important challenge,
making direct methods computationally expensive and preventing its use for
online motion generation. This fact is accentuated by including the contact
forces as decision variables. In \cite{Dai2014} the use of the simpler, yet
completely valid, centroidal dynamics of the robot was proposed to overcome this
difficulty. 
The length of the problem is then reduced with respect to the one using the full
model. Nevertheless, using the centroidal dynamics prevents the use of 
constraints or costs based on the joint torques, moreover additional kinematic 
constraints are required to obtain feasible solutions.

Another technique known as multiple shooting combines the shooting and direct 
approaches. Multiple shooting has been used for human-like whole body optimal
control \cite{Schultz2010} and to the analysis of specific motion
problems on humanoids \cite{Koch2014}. Similar to direct methods, constraints 
can be explicitly formulated. As in shooting methods, however, this approach
needs an explicit model for the contact dynamics to describe the changes between
contact configurations or phases.

In this paper we present a new approach for whole body motion planning based on
a direct method. Here we use a reduced version of the full body dynamics
obtained from the projection of the equations of motion into the tangent space
with respect to the constrained manifold. As a consequence of the projection,
contact forces are removed from the dynamics whereas its effects are still taken
into account. The main contribution of this paper is the integration of a direct
transcription method for trajectory optimization with a model representing the
direct dynamics of a legged robot where contact forces are not explicit. The
projection of the equations of motion is obtained via an algebraic operator that
depends only on kinematic quatities and thus obtaining solutions in a tractable
time. Moreover, compared with shooting methods, our approch does not required
an explicit contact model, reducing the complexity of the numerical optimization
process.

This paper is organized as follows. Section \ref{sec:Framework}
presents the optimal control problem of legged robots in a very general form 
and describes our solution approach.
Section \ref{sec:DirectDynamics} reviews the main concepts of the projection
of the equations of motion into a constraint space. In Section
\ref{sec:DirectTranscription} we describe the direct transcription 
method to generate motions in legged robots based on the projected model. 
Results are 
described in section \ref{sec:Results}. Finally, conclusions and future work are
discussed in section \ref{sec:Conclusions}.
 
\section{General Framework}
\label{sec:Framework}
In general, the system dynamics of a robot can be modelled by a set of
nonlinear differential equations,
\begin{equation}
\dot{\mathbf{x}}(t) = f(\mathbf{x}(t),\mathbf{u}(t)),
\label{eq:systemdynamics}
\end{equation}
where $\mathbf{x} \in \mathbb{R}^{n_x}$ represents the system states and
$\mathbf{u} \in \mathbb{R}^{n_u}$ the vector of inputs. The transition function
$f(\cdot)$ defines the system evolution over time.
Legged robots can be modeled as underactuated, 
floating base systems, i.e., the state space is given by
$\mathbf{x}=[\mathbf{q}^T,\dot{\mathbf{q}}^T]^T$ 
where the
generalized coordinate  $\mathbf{q = { [ q}^T_{\mathbf{b}} \quad
\mathbf{q}^T_{\mathbf{r}} ] }^ T$ includes base positions and orientations
$\left(\mathbf{q_b} \in \mathbb{R}^6\right)$ as well as joint configurations
$\left(\mathbf{q_r} \in \mathbb{R}^n \right)$, i.e., $n_x = 2\times(6+n)$. Control inputs are usually joint torques $\mathbf{u} = \tau \in \mathbb{R}^n$.

In a very general formulation, a trajectory optimization problem for whole body
motion planning and control in legged robots,

\begin{equation}
\arraycolsep=1.4pt\def\arraystretch{1.5}
\begin{array}{cc}
\underset{\mathbf{u}(t),\mathbf{x}(t)} {\text{min}} & J = \Psi(\mathbf{x}(t_f)) + \displaystyle \int_{0}^{t_f} \psi(\mathbf{x}(t),\mathbf{u}(t),t)~dt \\
\text{s.t.,} 
& \dot{\mathbf{x}} = f(\mathbf{x},\mathbf{u}) \quad \quad \\
& \mathbf{g}_{0,l} \leq \mathbf{g}_0(\mathbf{x}(t_0),\mathbf{u}(t_0),t_0) \leq \mathbf{g}_{0,u} \\
& \mathbf{g}_{f,l} \leq \mathbf{g}_f(\mathbf{x}(t_f),\mathbf{u}(t_f),t_f) \leq \mathbf{g}_{f,u} \\
& \mathbf{\phi}_{l} \leq \mathbf{\phi}(\mathbf{x}(t),\mathbf{u}(t),t) \leq \mathbf{\phi}_u \\
& \mathbf{x}_{min} \leq \mathbf{x}(t) \leq \mathbf{x}_{max} \quad \mathbf{u}_{min} \leq \mathbf{u}(t) \leq \mathbf{u}_{max}
\end{array}
\label{eq:to_problem}
\end{equation}
consists in finding a finite-time state and input trajectory $\mathbf{x}(t) , 
\mathbf{u}(t), \forall t \in [0,t_f]$, such that a given criteria $J$ is
minimized, subject to a set of constraints. Intermediate $\psi(\cdot)$ and final
$\Psi(\cdot)$ costs encode the objective of the task through $J$. The optimization
is subject to the dynamics of the system, as well as to boundary conditions,
$\mathbf{g}(\cdot)$, path constraints $\mathbf{\phi}(\cdot)$, and simple bounds
on the state and control variables.

The complexity of this optimization problem stems from the complexity
of the dynamics of a legged robot. Besides the underactuation, 
legged robots are constantly establishing and breaking contacts with the
environment adding constraints to its motion
and generating ground reaction forces.




%
In \cite{Aghili2005}, a method to derive direct dynamics of constrained
mechanical systems based on the notion of a projector operator is presented.
Constraint reaction forces are eliminated by projecting the original dynamics
equations into the tangent space with respect to the constraint manifold.
Moreover, the equations of motions are derived in a form that explicitly
relates the generalized input force to the acceleration.
The following section revisits the derivation of the
constraint-consistent dynamics, 
describing those
concepts relevant to understand our approach for planning in such a
subspace.

\section{Direct Dynamics of a legged robot}
\label{sec:DirectDynamics}

In general, the  dynamics of a legged robot can be modeled as a 
constrained multibody system, i.e., 
\begin{equation}
\mathbf{
M(q)\ddot{q} + h(q,\dot{q}) = S}^T \mathbf{\tau + F}_c
\label{eq:full_dynamics}
\end{equation}
subject to $m$ kinematic constraints,
\begin{equation}
\mathbf{\Phi(q) = 0}. 
\label{eq:kin_constraints}
\end{equation}
where $\mathbf{M} \in \mathbb{R}^{(n+6)
\times (n+6)}$ represents the inertia matrix, $\mathbf{h}
\in \mathbb{R}^{n+6}$ is a generalized force vector, gathering Gravitational,
friction, Coriolis and centrifugal effects. 
$\mathbb{\tau} \in \mathbb{R}^{n}$ is the vector of joint torques and
$\mathbf{S} \in \mathbb{R}^{n \times (n+6)}$ is the joint selection matrix that
reflects the underactuation. 
$\mathbf{F}_c \in \mathbb{R}^{n+6}$,  represents the generalized constraint
force acting on the robot DOF.

The set of constraints described in (\ref{eq:kin_constraints}) models that 
there are certain points, $c_j$, of the robot which motion is instantaneously
constrained, and their velocity is zero, i.e., 
\begin{equation}
\mathbf{J}_c \dot{\mathbf{q}} = \mathbf{0}
\label{eq:velocity_constraint}
\end{equation}
where $\mathbf{J}_c = \partial \mathbf{\Phi} / \partial \mathbf{q} \in
\mathbb{R}^{m \times (n+6) }$  is the Jacobian of the constraints with
respect to the generalized coordinate. This can be used to express the
generalized constraint force,
\begin{equation}
\mathbf{F}_c = \mathbf{J}^T_c \mathbf{\lambda}
\end{equation}
where $\mathbf{\lambda} \in \mathbb{R}^m$ are the so-called Lagrange multipliers
of the constraints. Notice that no assumption has been made on the rank of
(\ref{eq:kin_constraints}), and it may contain redundancies. For instance, in
case of a point feet quadruped robot with all feet in contact with the ground,
$\mathbf{\lambda} \in \mathbb{R}^{12}$ represents the ground reaction forces on
the feet, hence the constraints in (\ref{eq:kin_constraints}) are redundant and
$\mathbf{\Phi}(\mathbf{q})$ is not full rank.


As first suggested in \cite{Aghili2005}, constraint forces can be eliminated from
(\ref{eq:full_dynamics}) by projecting the dynamics onto the null space of
$\mathbf{J}_c$, i.e., onto the tangent space with respect to the constrained
manifold.
Provided an orthogonal projection operator $\mathbf{P}$ such that its range is
equal to the null space of the Jacobian, i.e., $\mathcal{R}(\mathbf{P}) =
\mathcal{N}(\mathbf{J}_c)$, it can be shown that such operator is an annihilator
for the generalized constraint force\footnote{\label{note1}As shown in 
\cite{Aghili2005}, giving that $\mathbf{F}_c \in \mathcal{R}(\mathbf{J}_c^T)$ and by
virtue of the fundamental relationship between the range space and the null
space orthogonal associated with a linear operator and its transpose, it can be
shown that $\mathbf{F}_c \in \mathcal{N}(\mathbf{J}_c)^{\perp}$. Vectors can be
expressed as the sum of orthogonal components, $\dot{\mathbf{q}} = \dot{\mathbf{q}}_{\|}
\oplus \dot{\mathbf{q}}_{\perp}$. According to (\ref{eq:velocity_constraint}), $\dot{\mathbf{q}}\in \mathcal{N}(\mathbf{J}_c)$,
and as $\dot{\mathbf{q}}_{\|} = \mathbf{P}\dot{\mathbf{q}}$, i.e., $\dot{\mathbf{q}}_{\|}  \in \mathcal{N}(\mathbf{J}_c)$, therefore,  
$\dot{\mathbf{q}}_{\perp} = 0$.}, 
i.e., $\mathbf{PF}_c = 0$.
Applying the orthogonal operator to (\ref{eq:full_dynamics}),
the projected inverse dynamics of a constrained multibody system is then expressed in a descriptive form by,
\begin{equation}
\mathbf{PM\ddot{q}} = \mathbf{P(S}^{T}\tau - \mathbf{h})
\label{eq:projected_dynamics}
\end{equation}

This equation could already be used as a constraint for the optimization,
as it does not include the contact forces and the complexity of the
original problem would be reduced. However, this expression does not consider
the components of the null space orthogonal to the acceleration, i.e., the
component of acceleration produced exclusively by the constraint and not by
dynamics. The velocity constraint in (\ref{eq:velocity_constraint}) implies that
the null space orthogonal component of the velocity 
must be equal to zero.

\begin{equation}
\dot{\mathbf{q}}_{\perp} = \mathbf{(I-P)\dot{q}} = \mathbf{0}.
\label{eq:perp_qd}
\end{equation}
Differentiating (\ref{eq:perp_qd})
with respect to time, it can be seen that the null space orthogonal component
of the acceleration $(\ddot{\mathbf{q}}_{\perp})$ is not necessarily equal to
zero,
\begin{equation}
\ddot{\mathbf{q}}_{\perp} = \mathbf{(I-P)\ddot{q}} = \mathbf{C\dot{q}}
\label{eq:perp_qdd}
\end{equation}
where $\mathbf{C} = d\mathbf{P}/dt$. 
Aghili proposed in \cite{Aghili2005} that  combining (\ref{eq:projected_dynamics}) and
(\ref{eq:perp_qdd}), and assuming a square projector matrix $\mathbf{P} \in
\mathbb{R}^{(n+6)\times(n+6)}$, a complete equation of motion of the constrained
mechanical system can be obtained,
\begin{equation}
\ddot{\mathbf{q}} = \mathbf{M}_c^{-1} \left(\mathbf{S}^T\tau - \mathbf{h} + \mathbf{C}_c \dot{\mathbf{q}} \right)
\label{eq:direct_proj_dynamics}
\end{equation}
where $\mathbf{C}_c = \mathbf{MC}$, whereas $\mathbf{M}_c \in \mathbb{R}^{(n+6)\times(n+6)}$,
the so-called \emph{constraint inertia matrix}, is defined as,
\begin{equation}
\begin{array}{lcl}
\mathbf{M}_c  &  =  & \mathbf{M} + \tilde{\mathbf{M}} \\
\tilde{\mathbf{M}} &  =  & \mathbf{PM} - (\mathbf{PM})^T.
\end{array}
\label{eq:constraint_inertia_matrix}
\end{equation}

The equation of motion in (\ref{eq:direct_proj_dynamics}) is completely
compatible with the trajectory optimization problem in  (\ref{eq:to_problem})
and with the direct transcription approach described in the next section.
It is worth mentioning that, as shown in \cite{Aghili2005},
there is no unique way of combining (\ref{eq:projected_dynamics}) and
(\ref{eq:perp_qdd}) to obtain (\ref{eq:direct_proj_dynamics}). Although all
forms are equivalent, some might have computational advantages over others.
As demonstrated in \cite{Aghili2005}, the constraint inertia matrix 
in (\ref{eq:constraint_inertia_matrix}) is positive definite but not
necessarily symmetric.

Finally, it is important to remark that the derivation of
(\ref{eq:direct_proj_dynamics}) assumes a square projector. That is the case of
the orthogonal projector proposed in \cite{Aghili2005}, and used throughout this
paper, 
\begin{equation}
\mathbf{P = I - J}_c^{+}\mathbf{J}_c
\end{equation}
where $\mathbf{J_c^{+}}$ denotes the pseudoinverse of $\mathbf{J}_c$. Other
forms of projector operators have been proposed in the context of inverse
dynamics contol of floating base robots \cite{Righetti2011}. Nevertheless, 
these are not square 
and therefore cannot be used to compute a direct dynamics expression compatible with the
direct transcription approach used in this work.
\section{Direct Transcription}
\label{sec:DirectTranscription}
We use direct transcription \cite{Betts1998,Hargraves1987} to find a feasible motion plan. This method
translates the continuous formulation in (\ref{eq:to_problem}) into a
mathematical optimization problem with a finite number of variables that can be
solved using an Nonlinear Programming Solver (NLP).

The decision variables $\mathbf{y} \in \mathbb{R}^p$ of the transcription
problem are the discrete values of the robot state  and control trajectories
sampled at certain points or nodes. The time between nodes $\Delta T$ is not 
necessarily constant and has been included in the set of decision variables,
i.e., $\mathbf{y} =\left\{\mathbf{x}_k,\mathbf{u}_k,\Delta T_k\right\}$ for $k =
1,...,N$, where $N$ represents the total number of nodes.

The resulting NLP is then formulated as follows,
\begin{equation}
	\begin{array}{rlclcl}
		\displaystyle \min_{\mathbf{y}} & f_0(\mathbf{y}) \\
		\textrm{s.t.}, & & \mathbf{\zeta}(\mathbf{y}) & =&  0& \\ 
		&  \mathbf{b}_{min}  &\leq &\mathbf{b}(\mathbf{y})& \leq & \mathbf{b}_{max} \\
		& \mathbf{y}_{min}  &\leq &\mathbf{y}& \leq  &\mathbf{y}_{max}, \\		
	\end{array}
\label{eq:NLP}
\end{equation}
which contains a scalar and derivable objective function $f_0(\mathbf{y})$, a set of boundary and
path constraints $\mathbf{b}(\cdot)$ and bounds [$\mathbf{y}_{min}, \mathbf{y}_{max}$]
on the state and control variables.

\subsection{Dynamic constraints}

The original set of differential equations representing the dynamics of the
robot in (\ref{eq:systemdynamics}) is replaced with a vector of dynamic
constraints or \textsl{defects} $\mathbf{\zeta}(\cdot) \in
\mathbb{R}^{(N-1)\times n_x}$. Although there are different alternatives to 
formulate the dynamic constraints \cite{Pardo2015,Betts1998}, we use
a trapezoidal approximation for the dynamics, 
\begin{equation}
\mathbf{\zeta} \doteq \mathbf{x}_{k+1} - \mathbf{x}_k - \frac{\Delta T}{2} \left[ f(t_k) + f(t_{k+1}) \right] = 0
\end{equation}
where the notation $f(t_i) = f(\mathbf{x}_i,\mathbf{u}_i)$ has been adopted for
simplicity. Notice that the inverse dynamics in (\ref{eq:projected_dynamics})
cannot be used to compute this form of defect as $f(\cdot)$ cannot be
computed through matrix inversion. On the other hand, the defects can be easily
computed using (\ref{eq:direct_proj_dynamics}).

\subsection{Torque bounds as path constraints}

\begin{figure}
	\centering
	\includegraphics[width=0.75\columnwidth, keepaspectratio=true]{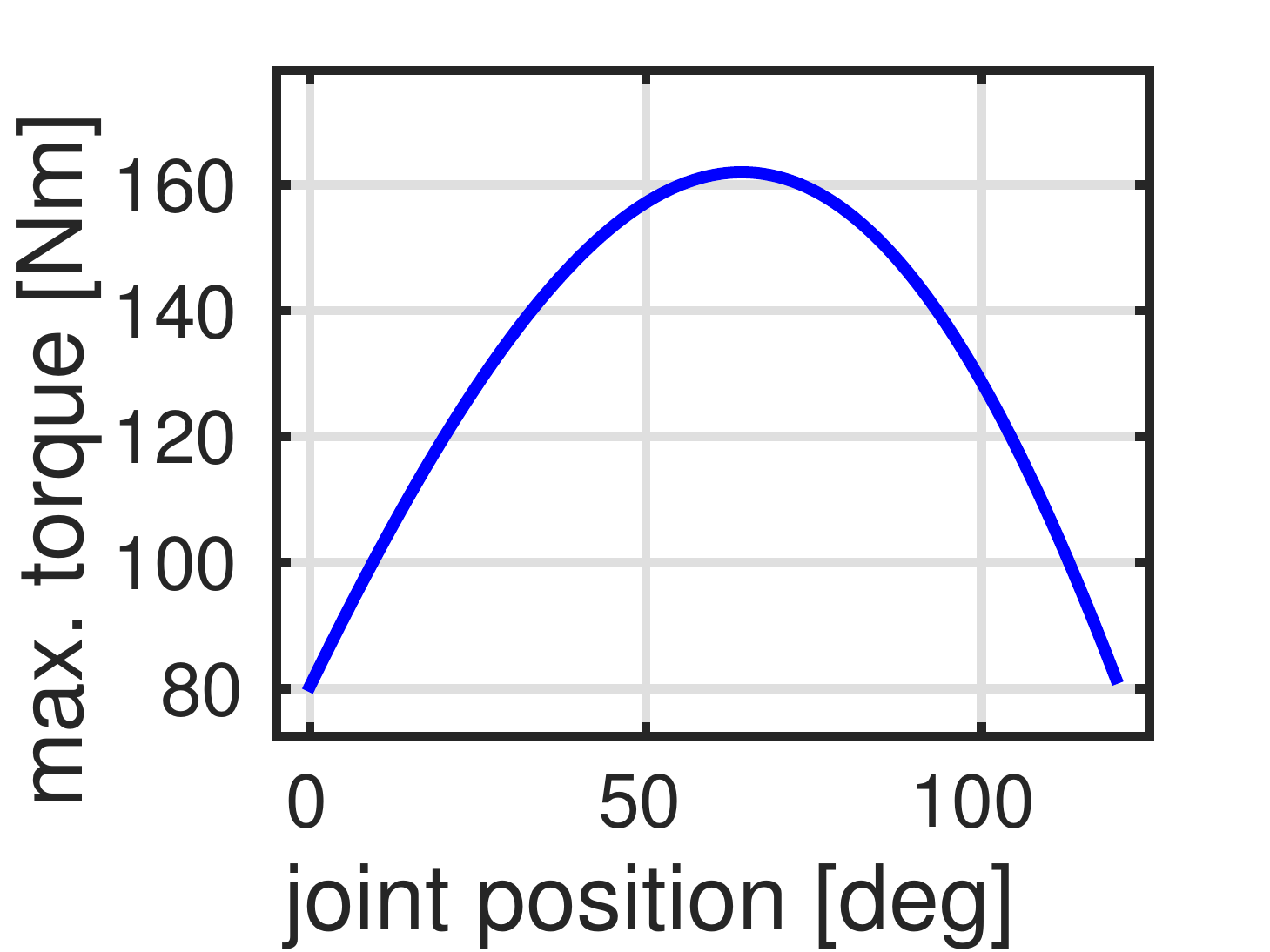}
	\caption{Maximum torque of HyQ knee 	
	is function of the joint configuration due to a varying leverarm. This function is then used as path constraint to avoid limiting the capacities of the robot.}
\label{fig:lever_dependent_tau}
\end{figure}

As in the case of hydraulically actuated robots, torque limits may depend on
the configuration of the robot, i.e.,
$\tau_{max} = \varphi(\mathbf{q})$. For instance, Fig. \ref{fig:lever_dependent_tau}
shows the maximum torque that can be applied at the knee joint of the 
hydraulically-actuated quadruped (HyQ) \cite{Semini2011}.
We have added path constraints allowing
the robot to use the actual maximal torque available, and thus exploiting its dynamic
capacities. The constraint can be transcribed for the value of the torque at each node as follows,
\begin{equation}
0 \leq \varphi_i(\mathbf{q}) - \tau_i \leq \infty
\label{eq:torque_path_constraint}
\end{equation}
%

\subsection{Contact Points and Kinematic constraints}

It is important to add kinematic constraints to satisfy the assumption that
points $c_j$ are in contact, and therefore the projection is valid. In general,
if $\mathcal{S}_j$ groups the valid contact regions (floor, wall, etc), the contact point
used to generate the projection (i.e., $\mathbf{J}_c$) should satisfy,
$c_j \in \mathcal{S}_j.$

For instance, in the case of the quadruped robot used in this paper,
contact points are in the feet and the valid contact region is the ground,
therefore the contact point constraint can be expressed as,
\begin{equation}
f^{c_j}_{kin}(\mathbf{q})=0.
\end{equation}
where $f^{c_j}_{kin} \in \mathbb{R}$ is the kinematic function
providing the perpendicular distance from the foot to the ground.

In order to facilitate the numerical optimization, other
kinematic constraints that enforce the assumptions made to derive the projected dynamics
may be added. For instance, when the dynamic constraints are satisfied, those
kinematic constraints in (\ref{eq:velocity_constraint}) and its derivative, i.e.,
\begin{equation}
\mathbf{J\ddot{q}} -\mathbf{\dot{J}\dot{q}} = 0,
\label{eq:accel_constraint}
\end{equation}
are necessarily being satisfied. Adding this constraints is straightforward and
we observed that this helps the solver in finding feasible solutions.

\subsection{Friction cone constraint}

In order to generate feasible solutions, ground reaction forces at contact points $\lambda_{j} \in \mathbb{R}^3$ should
be unilateral (i.e., the robot only can push to the ground) and its tangential
component $\mathbf{\lambda}_j^{x,y}$ should 
avoid sliding and mantain the contact, i.e., respecting the friction cone. 
This constraint can be written as,

\begin{equation}
- \infty \leq ||\lambda_j^{x,y}|| - \mu \cdot \lambda_j^{z} \leq 0
\end{equation}
where $\mu$ represents the friction coeficient between the contact point and the surface. This constraint already imposes the unilateral condition. 

Even though contact forces are not part of the decision variables, 
it can be
uniquely defined as a function of the state and control by an algebraic
expression resulting from decoupling the 
dynamics, as derived in \cite{Righetti2011}. 
\begin{equation}
\lambda_i = \mathbf{R}^{-1}\mathbf{Q}_c^T(\mathbf{M} \cdot f(t_i) + \mathbf{h}- \mathbf{S}^{T}\tau _i)
\end{equation}
where $\mathbf{Q}_c$ and $\mathbf{R}$ derive from the QR decomposition of the constraint
Jacobian $\mathbf{J}_c^T = \mathbf{Q}[\mathbf{R}^T \ \mathbf{0}]^T$, and, $\mathbf{Q} = [\mathbf{Q}_c \ \mathbf{Q}_u]$.

\subsection{Switching contact configuration}

\begin{figure}
	\centering
	\includegraphics[width=0.75\columnwidth, keepaspectratio=true]{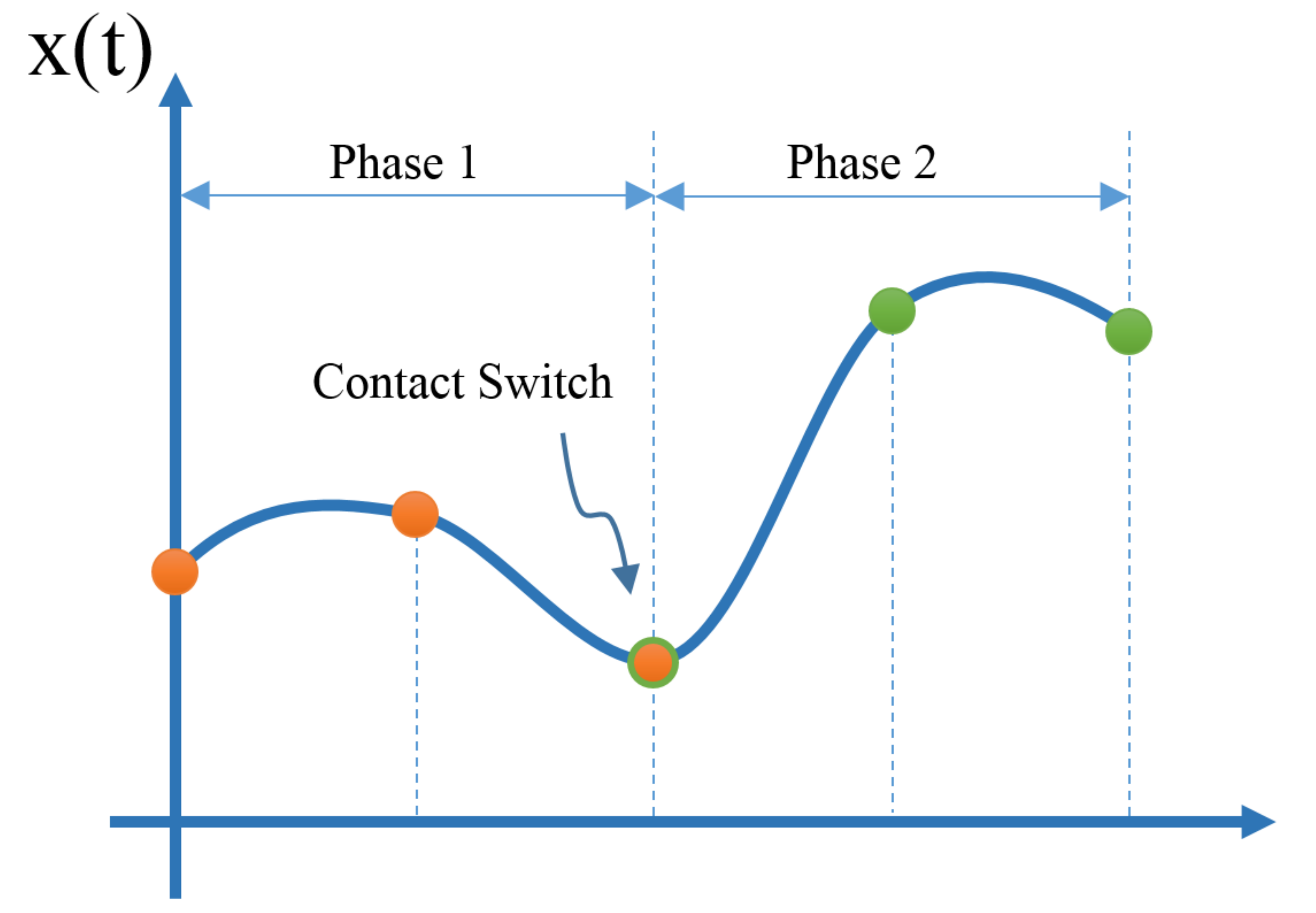}
	\caption{The phases of the optimization are defined by the contact configuration of the robot. Transition between phases is handled automatically by the method, as dynamic constraints are verified in between the nodes. Last point of a phase also satisfy the dynamic constraint of the next phase.}
\label{fig:phase_graph}
\end{figure}

Given that the dynamics are different for each contact configuration, the
sequence of contacts defines different optimization \emph{phases} \cite{Betts1998}.
The sequence of contact configuration has to be defined beforehand,
while the duration of the phases is optimized over. 
Transitions
between phases are handled automatically by the method, as constraints are
veryfied in between nodes. At the transition nodes the trajectory satisfy
the dynamics of both phases (see Fig. \ref{fig:phase_graph}).

\section{Results}
\label{sec:Results}

We have applied the framework described above to the generation of motions on the quadruped robot HyQ. This section describes the robot and shows the results obtained.

%

\subsection{HyQ Robot and Conventions}

\begin{figure}
	\centering
	\includegraphics[width=0.55\columnwidth, keepaspectratio=true]{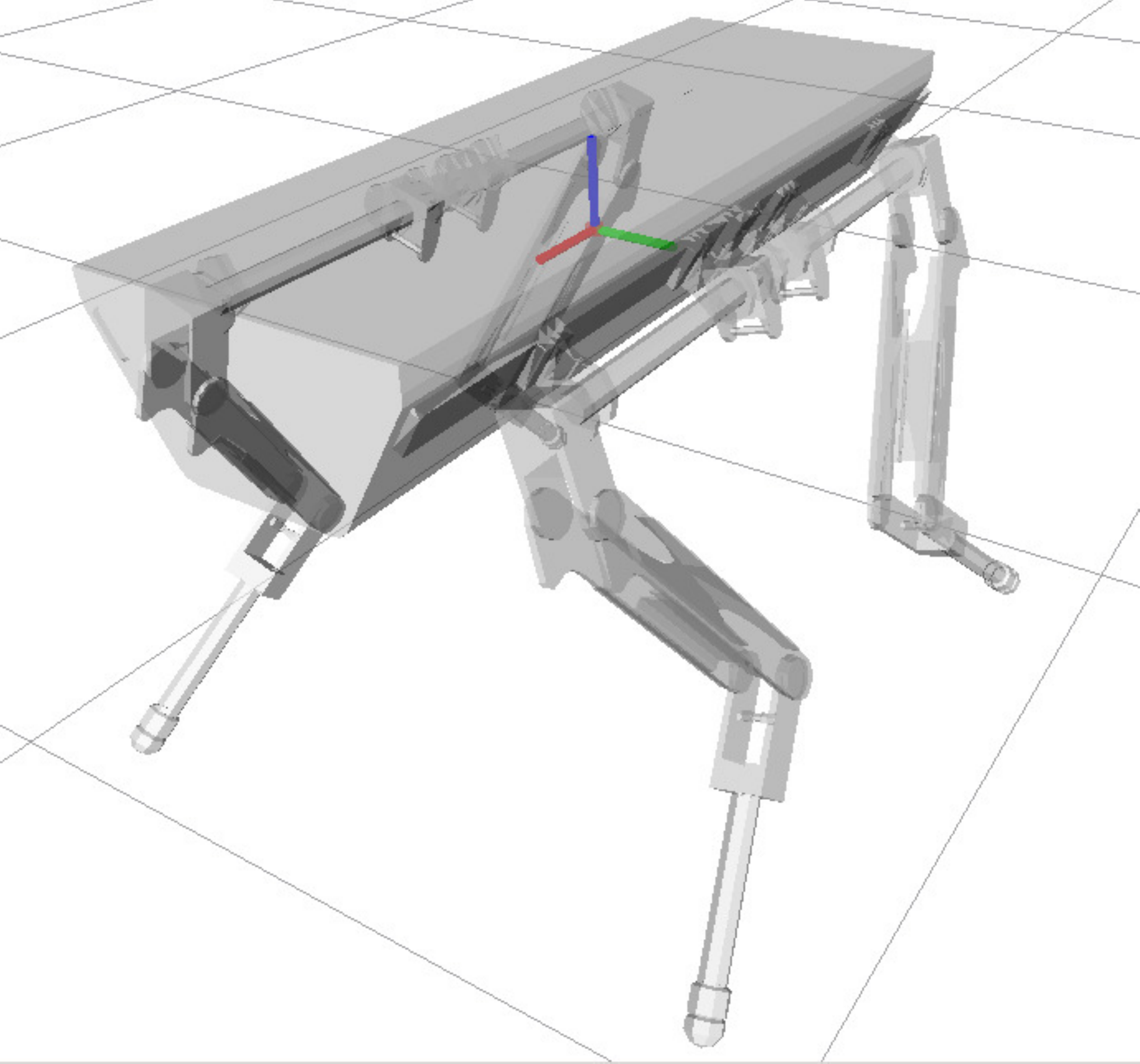}
	\caption{Unified Robot Description Format (URDF) model of the Hydraulically-actuated Quadruped (HyQ). Dimensions at nominal configuration : 80cmx40cmx60cm. Each leg has 3DOF (hip abduction, hip flexion and Knee) }
\label{fig:HyQRobot}
\end{figure}

Fig. \ref{fig:HyQRobot} shows the HyQ robot. This robot is approximately 80cm long and 50cm
wide, and it weights around 80Kg.  Each leg has three DOF (HAA: Hip  
abduction/adduction, HFE:
Hip flexion/extension and KFE: Knee flexion/extension - KFE) and it is 
fully torque controllable. Generating agile and
dynamic motions in this type of robots is challenging due to its rigid body dynamics.
At the same time the high  bandwith of hydraulic actuation offers
an opportunity to exploit dynamic capabilities.

The equations of motion of the robot are based on the rigid body dynamics model 
provided by \cite{Frigerio2012}, on top of this we implemented the
floating base dynamics using Plucker coordinates as suggested in
\cite{Featherstone2007}, i.e., linear and angular velocities of the base are
expressed in base coordinate frame, whereas position and orientation (XYZ Euler angles) are
represented with respect to an inertial frame.

\subsection{Feedback Stabilization}
\begin{figure}
	\centering
	\includegraphics[width=0.8\columnwidth, keepaspectratio=true]{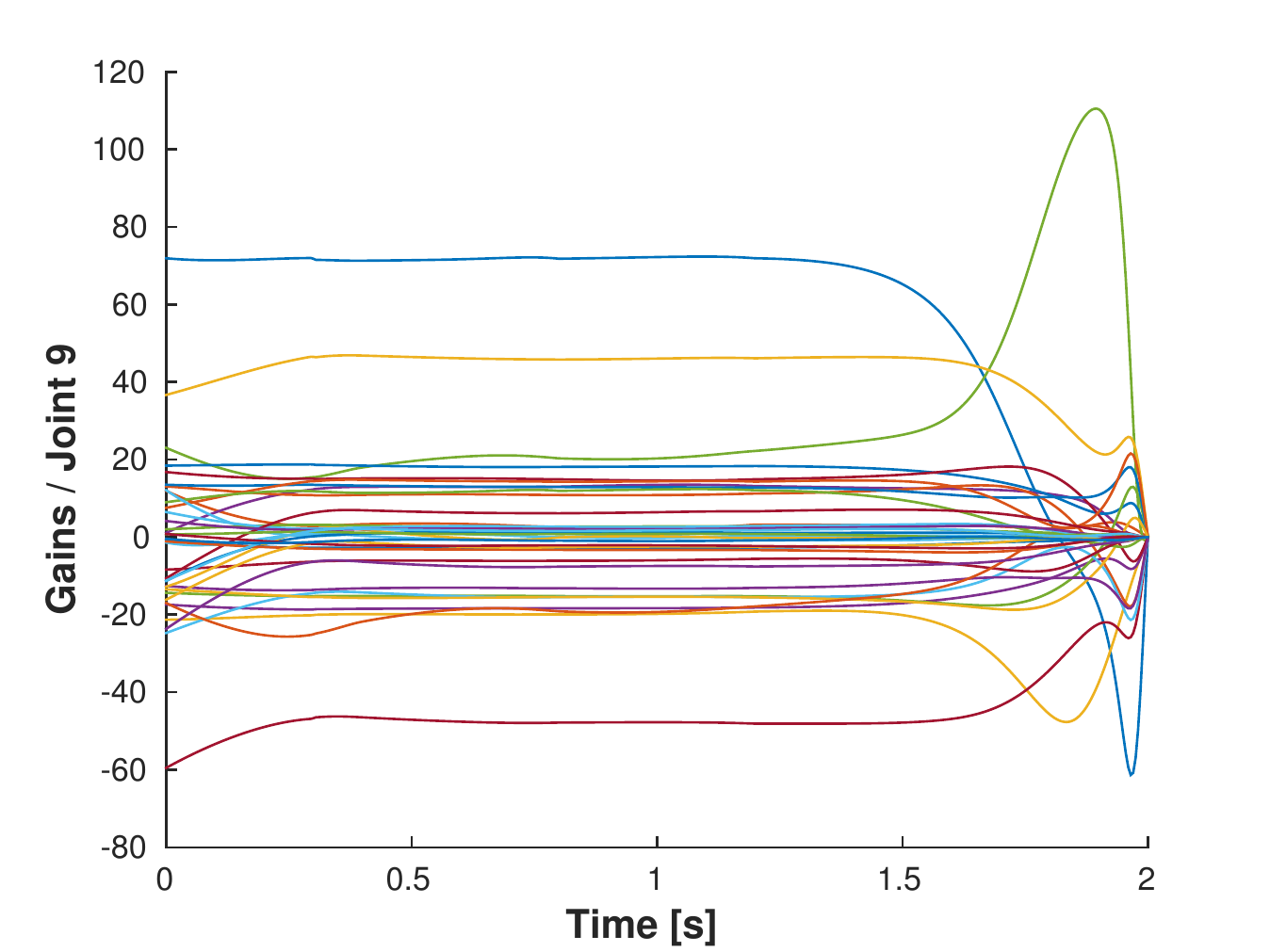}
	\caption{Gains Trajectories obtained using TVLQR. Whole body feedback is implemented and therefore all the state errors affects the final torque applied to the joints. This plot shows the gains corresponding to the feedback torque of Joint 9 (Left Hind Knee) during \emph{crouching} motion.}
\label{fig:Gains}
\end{figure}

In order to effectively apply the solutions of the optimization to the robot,
it is necessary to use a stabilizing feedback controller.
We use a Time Variant Linear Quadratic Regulator (TVLQR) \cite{Tedrake2014}
around
the optimized state and control trajectories $\mathbf{x}^*(t), \mathbf{u}^*(t)$.
This controller provides optimal full state feedback gains $\mathbf{K}(t) \in \mathbb{R}^{n_u\times n_x}$. The total control applied to the robot, $\mathbf{u}_T$, amounts to
$$
\mathbf{u}_T(t) = \mathbf{u}^*(t) + \mathbf{K}(t)( \mathbf{x}(t)-\mathbf{x}^*(t)). 
$$
Fig. \ref{fig:Gains} shows the trajectory of the 36 gains influencing the feedback control of a single torque command (e.g., Left Hind Knee) during a crouching motion.
We use the projected dynamics for the linearization and
subsequent solution of the differential Riccati equation. 

\subsection{Forward Integration and Simulation Environment}

Optimized trajectories were initialy tested performing the forward integration
of (\ref{eq:direct_proj_dynamics}). Nevertheless, this is a limited test as it
does not include contact reaction forces and it assumes perfect sensor
information of the pose of the robot body. Therefore, results were also
validated in a simulation environment \cite{Schaal2009} with an independent
contact and noise model. Finally, and in order to perform the validation under
the most realistic conditions, the feedback of the state of the base was
obtained using a state estimator.

\subsection{Single Phase Motions}

\begin{figure}
	\centering
	 \includegraphics[width=0.9\columnwidth, keepaspectratio=true]{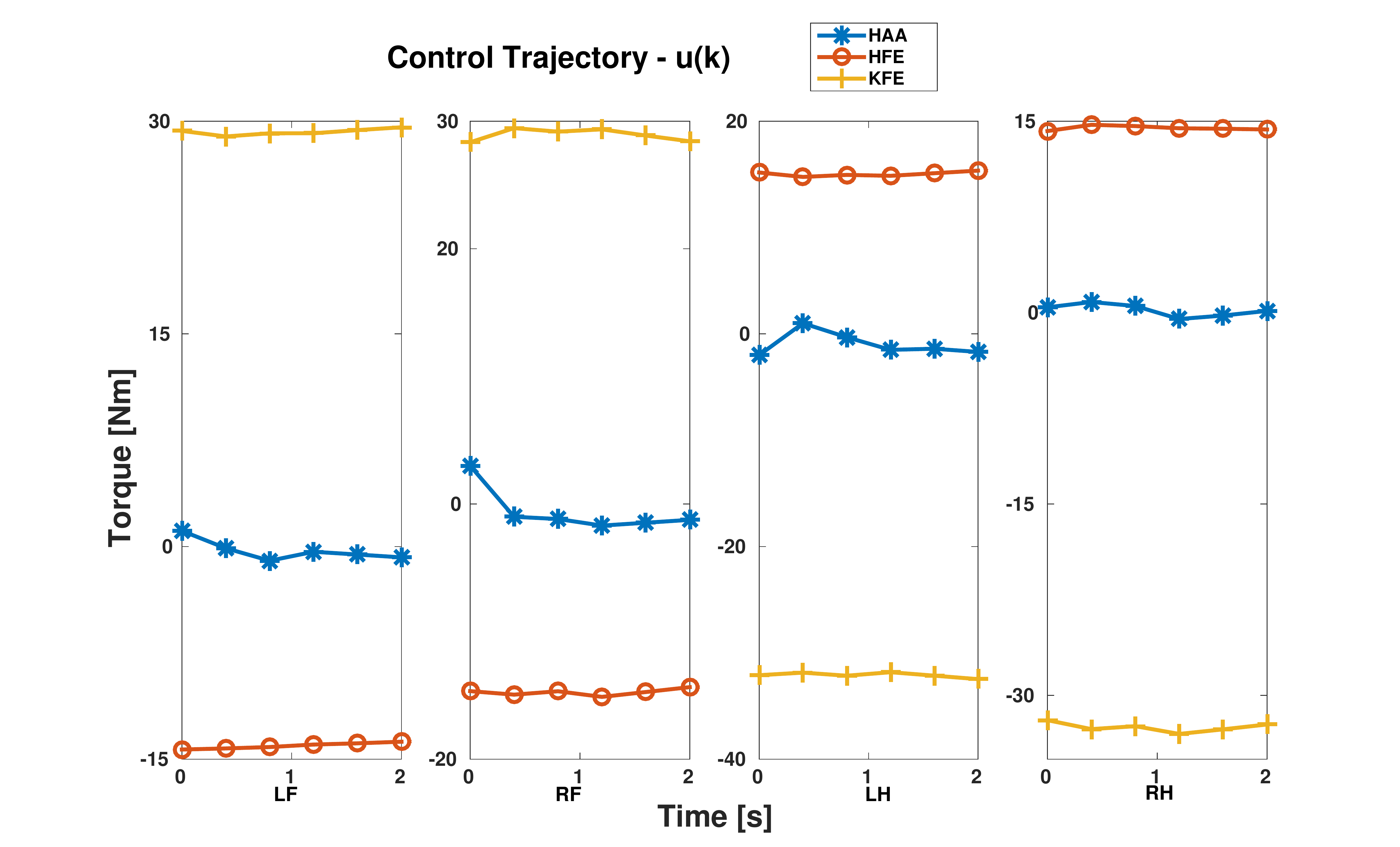}
	 \caption{Output of the trajectory optimization for the \emph{standing} example. Three torques trajectories per leg are shown : HAA (blue star), HFE (yellow plus) and KFE (red circles)}
\label{fig:Output}
\end{figure}

\begin{figure}
	\centering
	\includegraphics[width=\columnwidth, keepaspectratio=true]{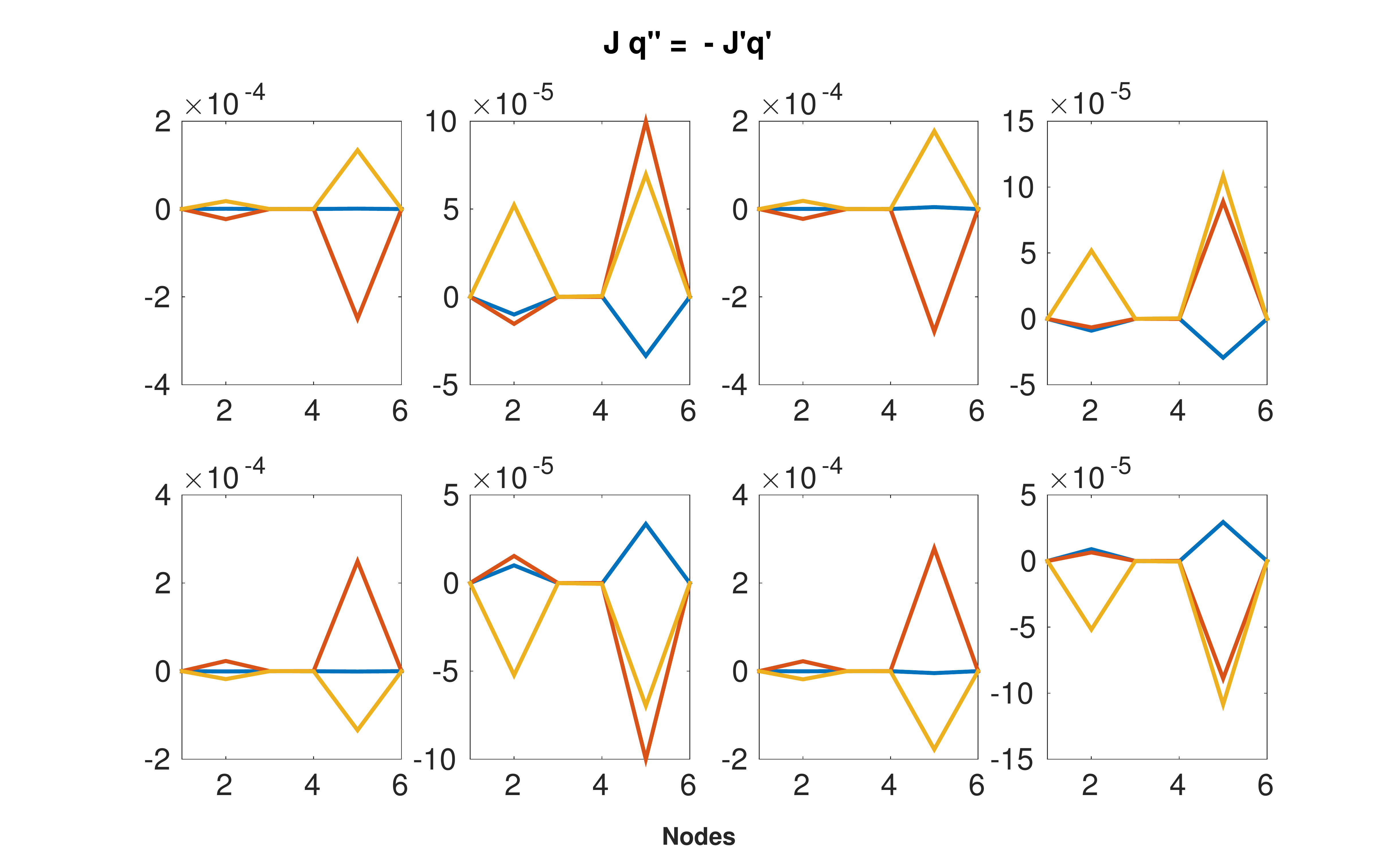}
	\caption{The first row of plots correspond to the term $\mathbf{J\ddot{q}}$,
	 for each leg at the discretization nodes. The second row correspond to the
	 term $\mathbf{\dot{J}\dot{q}}$. It can be observed that the second
	 derivative of the kinematic constraints is satisfied.}
\label{fig:dynamic_constraints}
\end{figure}

\begin{figure}[!ht]
	\centering
	\includegraphics[width=\columnwidth, keepaspectratio=true]{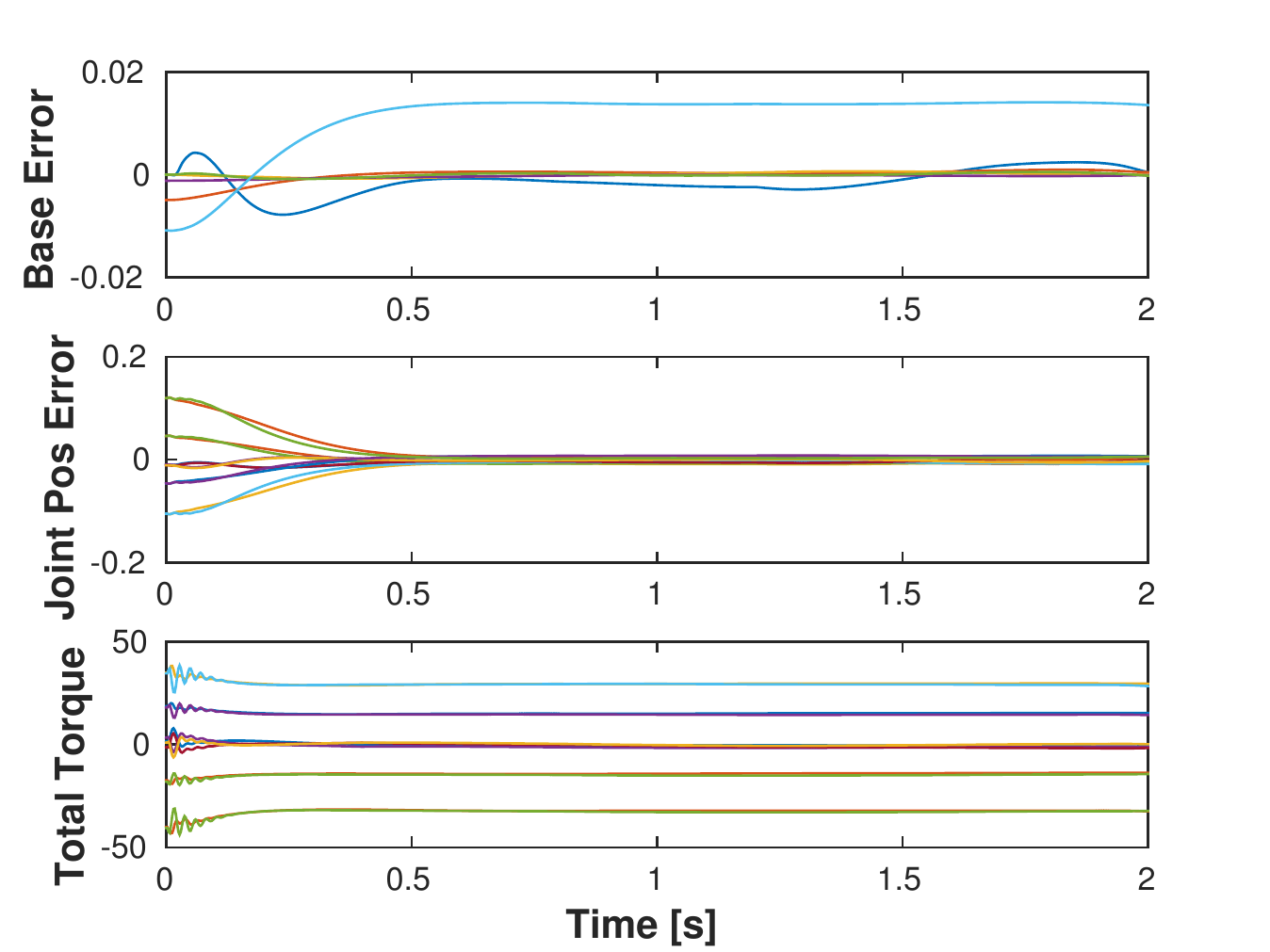}
	\caption{Simulation of a single phase motion with all feet on the ground. The
	cost function used for this motion penalizes any motion of the base, as a
	result, the robot stays standing. The plots shown here are intended to
	demonstrate the overall behavior of the robot during this task. Top : Base
	position and orientation error, stays bounded and near the origin. Middle :
	Joint position error of all joints converge to zero. Bottom: Torque commands
	are very low and the feedback component absorbes numerical integration errors
	as well as compensation for the effects of the contact model.}
\label{fig:Standing}
\end{figure}

Motions with single contact configuration were generated in order to validate all
the elements involved in the system. A quadratic cost function was used to 
obtain a \emph{standing} behavior
\begin{equation}
J_{standing} = \bar{\mathbf{x}}^T\mathbf{Q}_s\bar{\mathbf{x}} + \bar{\mathbf{u}}^T\mathbf{R}_s\bar{\mathbf{u}}
\end{equation}
where, $\bar{\mathbf{x}} = \mathbf{x} - \mathbf{x}_{nom}$ and $\bar{\mathbf{u}} = \mathbf{u} - \mathbf{u}_{nom}$ are the difference of the state and control with
respect to their nominal values. Nominal joint configuration for the standing
behavior is shown in Fig. \ref{fig:HyQRobot}. The base coordinate system and
the inertia coordinate system are aligned at the begininig of the mottion,
therefore the initial robot pose is given by $\mathbf{q_b = [ 0}_{1\times3} \
\mathbf{0}_{1\times3}]^T$.  Fig. \ref{fig:Output} shows the control signals
found for the optimization problem using $N=6$ discretization points for a
$T=2s$ trajectory. The contact constraint in (\ref{eq:accel_constraint}) is
verified by the signals in Fig. \ref{fig:dynamic_constraints}. 
Results of the simulation are shown in Fig. \ref{fig:Standing}.
These results confirm that control trajectories obtained using the projected 
dynamics are consistent with the dynamics of the robot since the complete system
follows the plan even though the simulation includes contact forces and noise. 

\begin{figure}[ht!]
	\centering
	\includegraphics[width=\columnwidth, keepaspectratio=true]{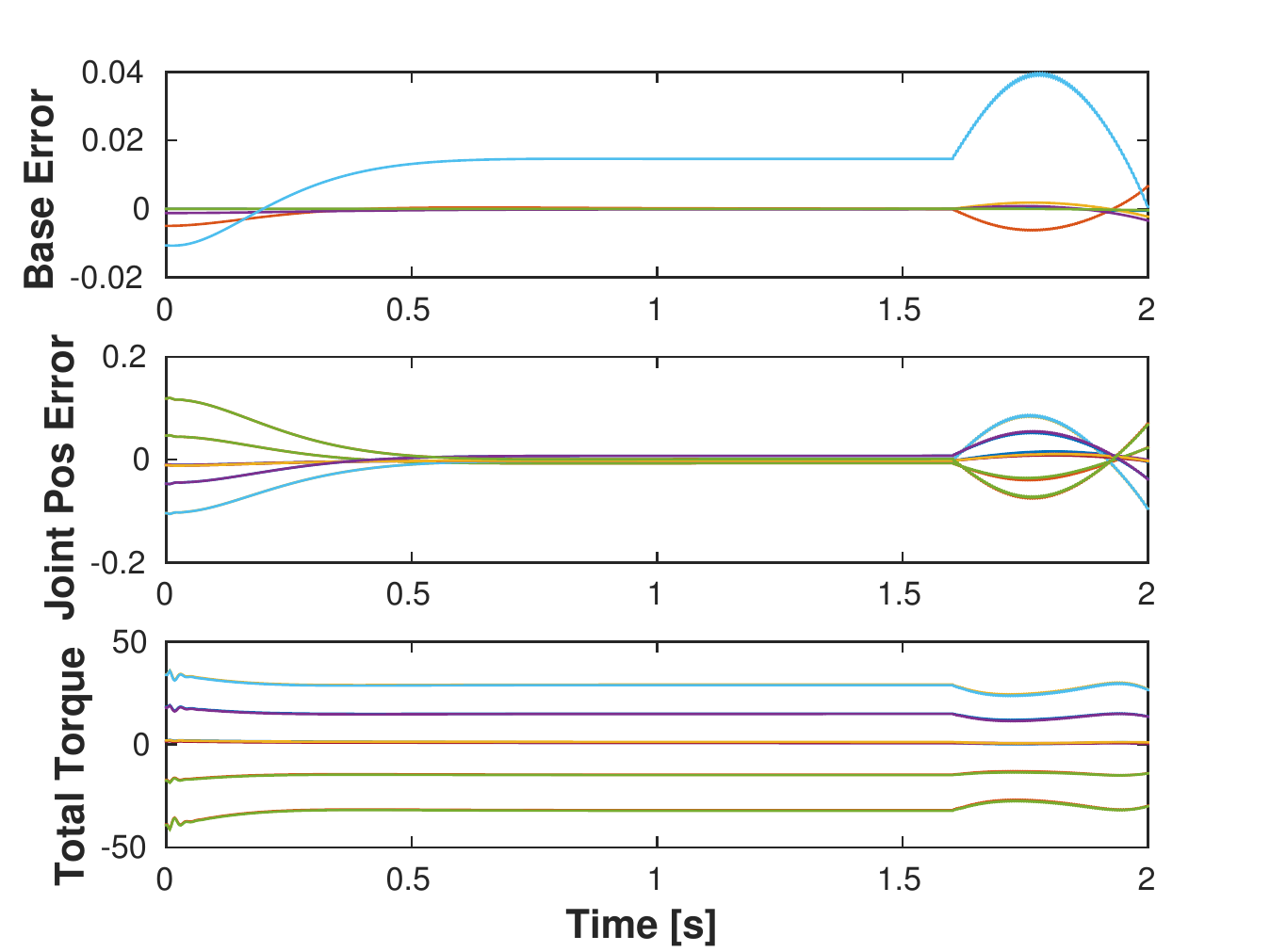}
	\caption{Simulation of a \emph{crouching} motion. Changing the cost and nominal
	value corresponding to the vertical component of the position of the base, the
	system generates trajectories for the states and controls in order to move the
	robot towards the desired position. Description of the plots is similar to the
	one used in Fig. \ref{fig:Standing}}.
\label{fig:Crouching}
\end{figure}

Modifying the nominal posture and cost corresponding to the $z$ component of the position of the base, the system finds trajectories and controls to crouch down to the desired position. Fig. \ref{fig:Crouching} shows the signals obtained during simulation. 

\subsection{Multiphase Motions}

\begin{figure}[ht]
    \centering
    \begin{minipage}[b]{.396\linewidth}
        \centering
        \includegraphics[width=\linewidth]{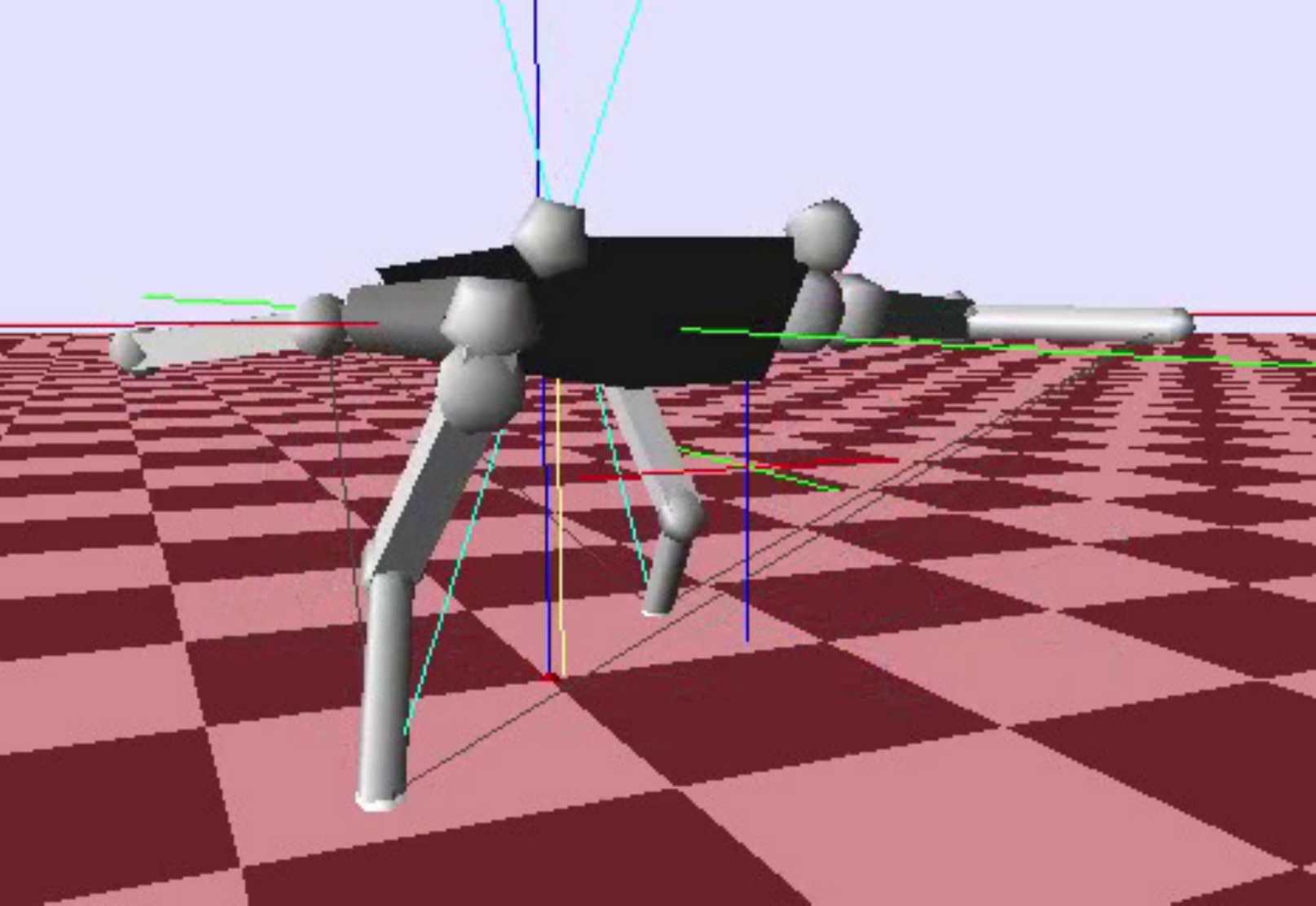}
        \caption{Balancing}
        \label{fig:stepping}
    \end{minipage}
    \begin{minipage}[b]{0.398\linewidth}
        \centering
        \includegraphics[width=\linewidth]{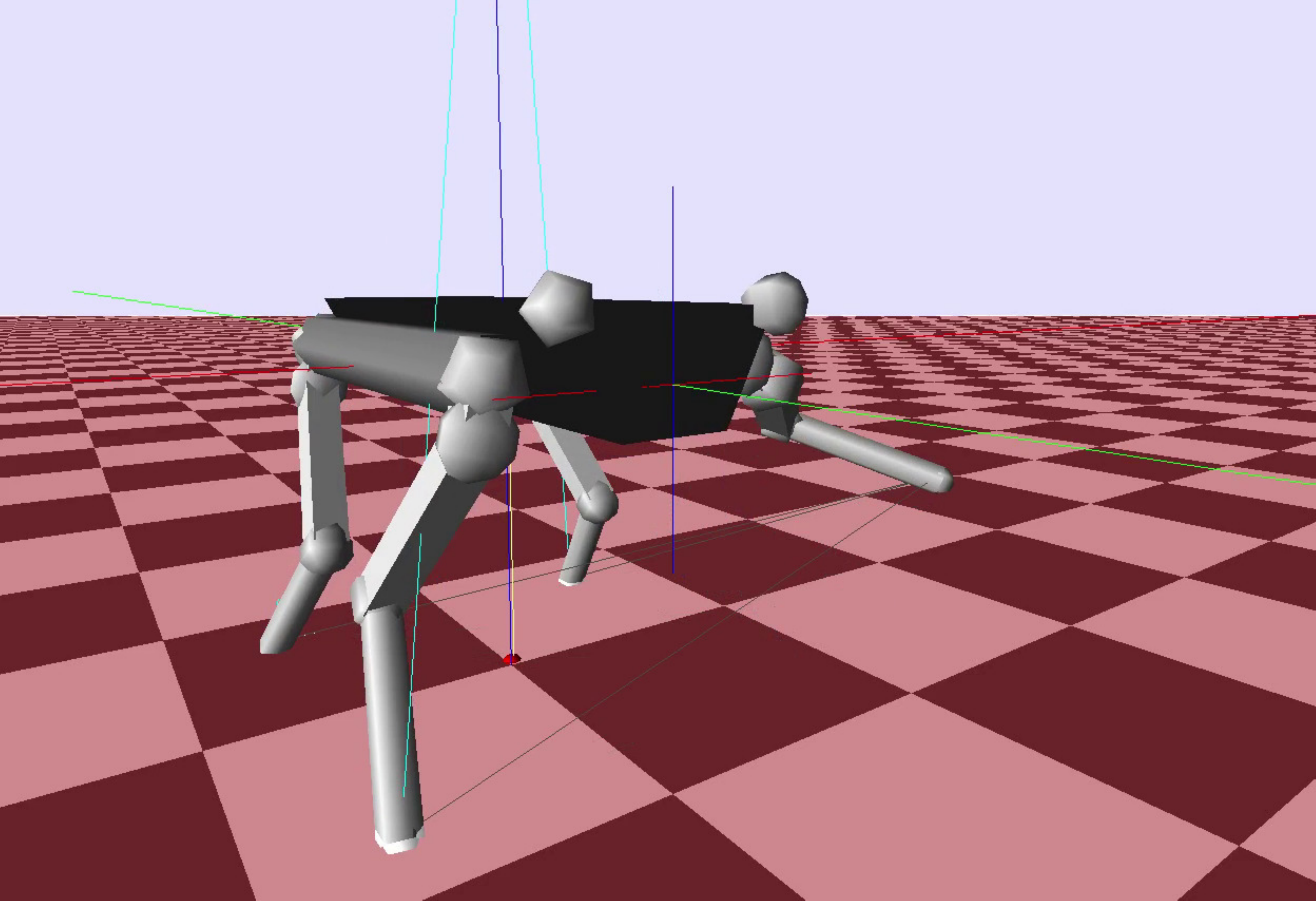}
        \caption{Stepping}
        \label{fig:rearing}
    \end{minipage}
    \begin{minipage}[b]{0.4\linewidth}
        \centering
        \includegraphics[width=\linewidth]{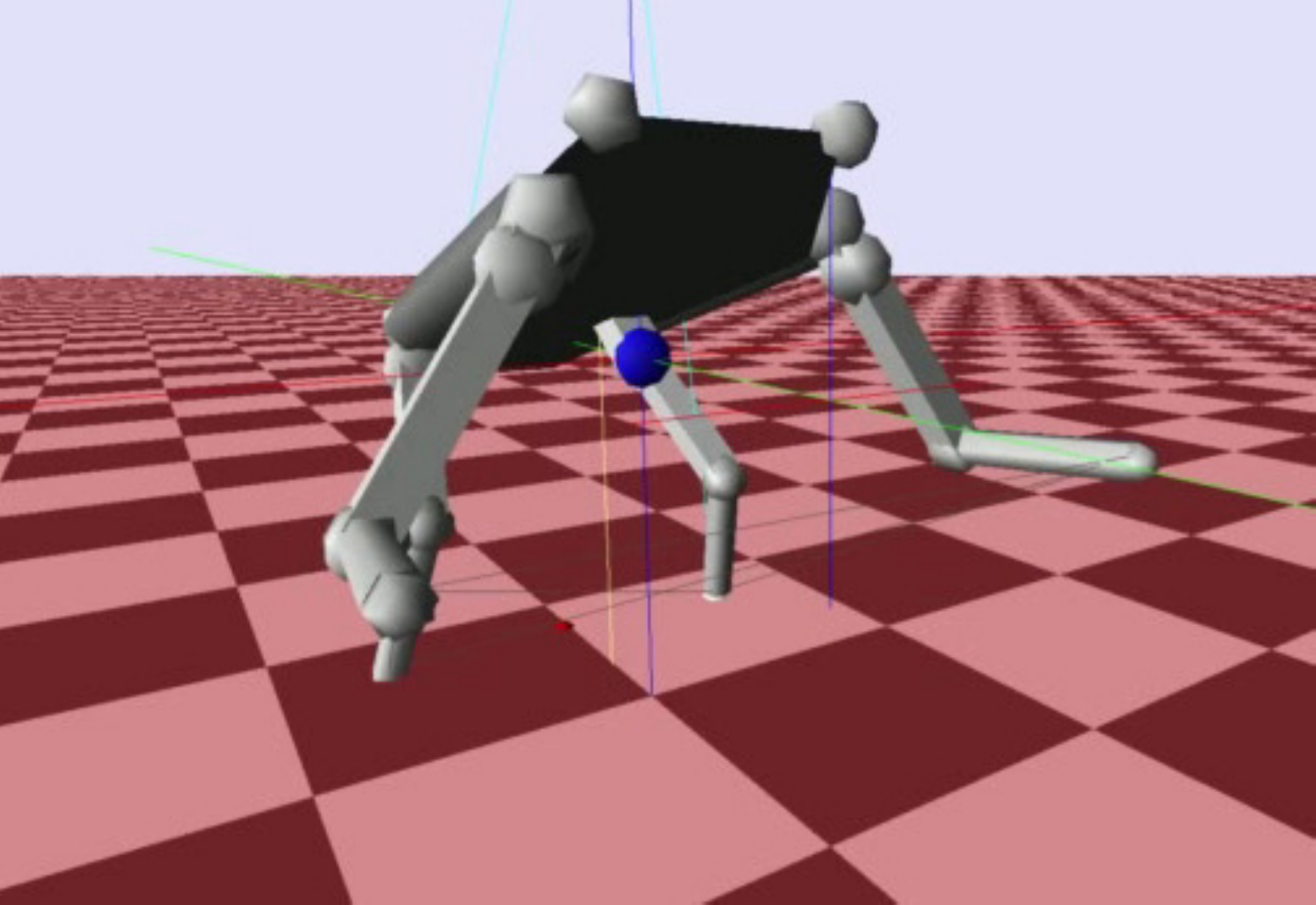}
        \caption{Rearing}
        \label{fig:balancing}
    \end{minipage}
\caption{Images captured in SL Simulation environment \cite{Schaal2009}. Final configuration corresponding to different multiphase motions (Stepping, Rearing , diagonal legs balancing).}
\label{fig:dynamic_motions}
\end{figure}

In order to generate more elaborated and dynamic motions, different
sequences of contact configurations were used.  Fig \ref{fig:dynamic_motions} shows the
final configuration corresponding to  the intermediate phase of different
motions (rearing, diagonal legs balancing and stepping). All motions start with the
same initial nominal posture and contact configuration.

The simulatuions described in this section are shown in the video attached to this paper:
\small{\url{https://youtu.be/_fy-E40evjE}}

\subsection{Comparison with centroidal dynamics}

All motions presented were obtained by the proposed optimization in few minutes
of processing in a common laptop. However, this is not a valid meassure of performance as too
many elements of the implementation are accountable to be optimized. 
In order to understand the impact of our contribution,
here we present a rough analyisis of the size of the problem compared with 
the state of the art on whole body motion planning using direct transcription 
methods \cite{Dai2014}.

Table \ref{tab:comparison} shows the total number of decision variables and 
constraints (per node) required in order to solve a single phase motion on
the HyQ robot (four feet on the ground)\footnote{As described in Section II-D of \cite{Dai2014} }. It can be seen that the complexity
of the problem is reduced using the projected dynamics and thus it can be 
hypothetized that the computational time required to solve the problem 
under same implementation  is less.

\begin{table}[h]
\caption{Number of decision variables and constraints (per node)
for a single phase motion on HyQ (all feet in contact)}
\label{tab:comparison}
\begin{center}
\begin{tabular}{|c||c|c|}
\hline
 Method & Decision Variables & Constraints\\
\hline
Centroidal Dynamics  & 76 & 63 \\
\hline
Projection  & 49 & 48\\
\hline
\end{tabular}
\end{center}
\end{table}

\subsection{Software Implementation}
Our system is developed in C++, based on the Eigen library for vector
manipulation and linear algebra. Given an implementation of the equation of
motion of the legged robot and a sequence of contact configurations, our systems
automatically projects the dynamics and generates the corresponding numerical
optimization problem. The system provides an interface to well known 
SNOPT \cite{Gill2005} solver, used to obtain the results presented in this paper.

\section{Conclusions and Future Work}
\label{sec:Conclusions}
The projection of the dynamics onto the null space of the Jacobians of its
contraints allows to reduce the complexity of the model, facilitating the use
of direct methods for trajectory optimization. The motions generated for the 
hydraulically-actuated robot satisfiy the kinematic constraints 
in (\ref{eq:velocity_constraint}) and (\ref{eq:accel_constraint})
even if they are not explicitly added to the optimization problem,
demonstrating the consistency of the formulation.
Motions including switching contacts (2 phases), have been shown to demonstrate
the feasibility of the method.

Single phase motions
were straightforward to 
implement in the real robot, however, we have observed that 
results tend to be at the limit of
dynamic stability, and therefore they are very sensitive to modelling errors and
noise. It seems that optimality of the solutions stress their feasiability (in the real robot).
Development of robust feedback controllers
is required to achieve the actual exploitation of the dynamics 
proposed by optimization approaches as the one presented in this paper.




\end{document}